\documentstyle[twocolumn]{article}
\leftmargin\leftmargini
\title{\Large \bf THE SPEECH-LANGUAGE INTERFACE IN THE \\ SPOKEN LANGUAGE
TRANSLATOR}

\author{\large David Carter and Manny Rayner\\
        \large SRI International \\
        \large Cambridge Computer Science Research Centre \\
        \large 23 Millers Yard \\
        \large Cambridge CB2 1RQ, U.K. \\
        \\
        \verb!dmc@cam.sri.com, manny@cam.sri.com!}

\date{\ }

\begin{document}

\maketitle

\section*{\centerline{\large \bf ABSTRACT}}

The Spoken Language Translator (SLT) is a prototype for practically
useful systems capable of translating continuous spoken language
within restricted domains. The prototype system translates air travel
(ATIS) queries from spoken English to spoken Swedish and to French. It
is constructed, with as few modifications as possible, from existing
pieces of speech and language processing software.

The speech recognizer and language understander are connected by a
fairly conventional pipelined N-best interface.  This paper focuses on
the ways in which the language processor makes intelligent use of the
sentence hypotheses delivered by the recognizer. These ways include
(1) producing modified hypotheses to reflect the possible presence of
repairs in the uttered word sequence; (2) fast parsing with a version
of the grammar automatically specialized to the more frequent
constructions in the training corpus; and (3) allowing syntactic and
semantic factors to interact with acoustic ones in the choice of a
meaning structure for translation, so that the acoustically preferred
hypothesis is not always selected even if it is within linguistic
coverage.

\section*{\large \bf \centerline{1. OVERVIEW OF}
\newline \centerline{THE SLT SYSTEM}}

The Spoken Language Translator (SLT) is a prototype system that
translates air travel (ATIS) queries from spoken English to spoken
Swedish and to French. It is constructed, with as few modifications as
possible, from existing pieces of speech and language processing
software.  This section gives a brief overview of the speech
recognition and language analysis parts of the SLT system and the
philosophy underlying them; for a longer treatment, including details
of transfer, generation, and speech synthesis, the reader is referred
to Agn\"{a}s {\it et al}, 1994. After the overview, we describe three
ways in which the language analyser makes intelligent use of the
N-best list of sentence hypotheses it receives from the recognizer.

At the highest level of generality, the design of SLT has two guiding
themes. The first is that of {\it intelligent reuse of standard
components}: most of the system is constructed from previously
existing pieces of software, which have been adapted for use in the
speech translation task with as few changes as possible. The second
theme is that of {\it robust interfacing}. In this paper, we focus on
an important means by which robustness is achieved: the delaying of
decisions about words, utterances and utterance meanings until
sufficient information is available to make those decisions reliably.

The speech recognizer used is a fast version of SRI's DECIPHER [TM]
speaker-independent continuous speech recognition system (Murveit {\it
et al}, 1991).  It uses context-dependent phonetic-based hidden Markov
models (HMMs) with discrete observation distributions for four
features: cepstrum, delta-cepstrum, energy and delta-energy. The
models are gender-independent and the system is trained on 19,000
sentences and has a 1381-word vocabulary. A bigram language model is
used. The output is an N-best hypothesis list, produced using a {\it
progressive recognition search} (Murveit {\it et al}, 1993) in which
the space of possible utterances is pruned by successively more
powerful but more costly techniques. The motivation for this kind of
search is to avoid making hard decisions without sufficient evidence,
while at the same time maintaining reasonable efficiency.

Fully-fledged linguistic analysis can be viewed from the perspective
of the speech recognition task as the final stage of progressive
search: the most powerful, most costly techniques used in the system,
exploiting complex syntactic and semantic knowledge, are applied,
reducing an already fairly limited set of possible utterances to a
single choice. Another, equally valid, perspective on language
analysis is from the standpoint of {\it utterance understanding}: the
purpose of source language processing in SLT is to map from the
acoustic signal to a representation of the utterance meaning, and
identifying the correct word sequence is a by-product of this process
rather than being a goal in its own right.

Language analysis in SLT is performed by the SRI Core Language Engine
(CLE), a general natural-language processing system developed at SRI
Cambridge (Alshawi, 1992).  The English grammar used for this is a
large general-purpose feature grammar, which has been augmented with a
small number of domain-specific rules.  It associates surface strings
with meaning representations in Quasi Logical Form (QLF; Alshawi and
Crouch, 1992).  Transfer and generation are performed by a second copy
of the CLE loaded with a French or Swedish grammar and transfer rules
for the appropriate language pair.

The system components are connected together in a pipelined sequence
as follows. First, DECIPHER processes the input signal and constructs
a list of N-best hypotheses, each tagged with an associated acoustic
score; N=5 gives a good tradeoff between speed and accuracy. The
construction of this list using the progressive search technique
constitutes a thorough pruning of the original search space of all
possible word sequences.

The hypothesis list is passed to the English-language version of the
CLE, which implements the final phase of progressive search by
applying the three processing stages outlined below and described more
fully in the remainder of this paper. The CLE achieves robustness in
the speech-language interface by postponing the selection of a correct
utterance (and utterance meaning) until all available knowledge has
been applied. This strategy is made acceptably efficient by the use of
a specialized fast parsing technique.  The processing stages are
these:

\begin{itemize}

\item As described in Section 2 below, the CLE examines the
hypotheses for evidence of speech repairs, and if it finds any, it
adds possible corrected versions to the list without removing the
originals, thus postponing a decision about whether the correction is
valid or not.

\item It then uses the grammar, specialized and compiled for both
speed and accuracy as described in Section 3, to analyze each
speech hypothesis (original and repaired) and extract a set of
possible QLF representations.  This typically results in a set of
between 5 and 50 QLFs per hypothesis.

\item The CLE's {\it preference component} is then used to
give each QLF a numerical score reflecting its {\it a priori}
linguistic (acoustic, syntactic, semantic and, within limits,
pragmatic) plausibility. The final score for a QLF is calculated as a
weighted sum of the scores assigned to it by a range of preference
functions, and the highest-scoring QLF is passed on for transfer and
target language generation. We describe the functioning of this
component in Section 4 below.

\end{itemize}

We now move on to examining these stages in more detail, starting with
the repair mechanism.

\section*{\large \bf \centerline{\large \bf 2. DETECTION AND} \newline
\centerline{CORRECTION OF REPAIRS}}

One important way in which spoken language differs from its written
counterpart is in the prevalence of self-repairs to speaker errors.
Examples such as the following occur in the ATIS domain:
\begin{enumerate}
\item {\small\tt list {\em LIST\/} FLIGHTS BETWEEN OAKLAND AND DENVER.}
\item {\small\tt does this {\em DOES THIS\/} FLIGHT SERVE \newline BREAKFAST.}
\item {\small\tt COULD I HAVE MORE DETAILS ON FLIGHT d l sixteen
{\em D L SEVEN\/} TWO SIX.}
\item {\small\tt SHOW ME ROUND TRIP FARES FOR flight two {\bf SORRY}
{\em FLIGHT FOUR\/} FOUR OH ZERO.}
\item {\small\tt I WANT A FLIGHT from boston {\em FROM DENVER TO BOSTON}.}
\item {\small\tt OK WHAT TYPES OF AIRCRAFT do {\em DOES\/} DELTA FLY.}
\end{enumerate}
In each case, the reparandum (material to be repaired) is shown in
{\small\tt lower case} and the repair itself in {\em ITALICS}, with
any explicit repair marker, such as ``sorry'', shown in {\bf BOLD}.
Note that, once the reparandum and any repair marker have been
identified, the location of the right hand end of the repair does not
affect the interpretation of the sentence (e.g.\ the repair in (3)
could be viewed as ``D L seven two six'').

In (1) and (2) the reparandum and repair are identical.  In (3)-(6)
they differ. (3) shows the substitution of a word after the repeated
material, (4) shows the use of an explicit repair marker, (5) is an
example of the additional material in the repair being inserted,
rather than appended, and (6) shows a correction of a suffix, with no
strictly identical words occurring.

However, not all repeated word sequences and (possible) explicit
repair markers indicate repairs; items (1') to (4') below are
non-repairs superficially similar to (1) to (4) above, with (5')
providing additional evidence that not all repetitions are repairs.
The typographic conventions show how the word sequences might be
misconstrued as repairs.
\begin{enumerate}
\item[1'.] {\small\tt SHOW ME ROUND TRIP FARES FOR U S FLIGHT four
{\em FOUR\/} oh {\em oh}.}
\item[2'.] {\small\tt IS u s {\em U S\/} AIR.}
\item[3'.] {\small\tt ARE ANY OF THE flights {\em NONSTOP FLIGHTS}.}
\item[4'.] {\small\tt I WANT a flight with {\bf NO} {\em STOPS}.}
\item[5'.] {\small\tt FROM PHILADELPHIA FROM DENVER AND FROM
PITTSBURGH.}
\end{enumerate}

It is known that repairs are often indicated acoustically (Bear {\it
et al}, 1992; Nakatani and Hirschberg, 1993) and DECIPHER could be
modified to detect possible repair indicators and pass the information
on to the CLE. However, this raises some difficult issues of
identification, representation and transportability, and it is worth
investigating how effectively repairs can be dealt with on the basis
of word strings alone.

In line with the philosophy behind progressive search, that of
postponing decisions until sufficient information is available, the
CLE's repair mechanism has the following novel feature: when a
possible repair is located, no immediate decision is made on whether
it is genuine. The (putatively) corrected word sequence is added to
the N-best list, and given a reduced acoustic score, without the
original hypothesis being removed.  Thus QLFs can be built from either
sequence, and the final choice of a word sequence is a by-product of
the choice of a QLF, which, just as for choices between original
hypotheses, takes advantage of full linguistic processing of all parts
of the sentence.

This methodology allows a range of repairs to be hypothesized by a
fairly straightforward algorithm while minimizing the number of false
positives found.  Given the word sequence actually uttered, it is in
general possible to determine the reality of a repair on the basis of
(a) specific, fairly superficial knowledge of what kinds of word
sequence tend (in the ATIS domain) to indicate repairs, (b) general
and ATIS-specific syntactic and semantic considerations, and (c)
knowledge of the discourse and reasoning about the domain. In the
translation task, a false positive --- ``correcting'' a non-existent
repair --- is a more serious error than failing to deal with a repair
that has occurred, because the former kind of error is likely to
confuse the user and to be viewed as much less acceptable.  The repair
detection algorithm therefore attempts to hypothesize just those
possible corrections that seem plausible on the basis of type (a)
knowledge and that, if they are false, are likely to be detectable
using type (b) knowledge, i.e.\ by syntactic, semantic and preference
processing.  Type (c) knowledge is not available within the SLT
system.

The detection mechanism identifies possible repairs by first searching for
repeated roots in the sentence, i.e.\ pairs of words (other than
numbers, which are often repeated intentionally) that can be analysed
morphologically by the CLE as having the same root. Examples are
``...flight...flight...'', ``...do...does...'' and ``...is...are...''.
It combines these pairs to identify sequences that begin and end
with the same roots, e.g.
\begin{quote}{\small\tt
I WANT TO GO \underline{FROM BOSTON} NO \underline{FROM DENVER TO
BOSTON} ON TUESDAY.
}\end{quote}
Sequences that have intervening material and consist only of one of a set
of very common words (``a'', ``and'', ``from'', ``in'', ``of'', ``or''
and ``to'') are discarded at this point, as inspection of the data
suggests they are likely to lead to false positives. In all other
cases, however, the two sequences (underlined above) are first matched
from left to right. Two points are awarded for a shared root, and one
is deducted when a word is skipped in either sequence. The match
proceeds (by dynamic programming) so as to maximize the score. In the
example, two points are awarded for the matches on each of ``from''
and ``Boston'' and one is deducted for skipping each of ``Denver'' and
``to''.

If there is no intervening material, the match is now complete, and
the hypothesized repair is returned. If there is intervening material
(as with ``no'' above) it may form part of either the repair or the
reparandum. Similar, but more general, matches are therefore carried
out in both the forward and the backward directions.

The forward match begins at the start of the intervening material and
just after the end of the second repetition sequence, i.e., at ``no''
and ``on'' in the example, and continues forwards until all the
intervening material is consumed. The backward match starts at the end
of the intervening material and just before the start of the first
sequence, i.e.\ at ``no'' and ``go'', and words backwards.  One point
is deducted for skipping a word in either sequence, unless the match
is forward and the word is known to be an explicit repair indicator,
in which case a point is awarded. (Explicit repairs are counted only
in the forward match to ensure they are identified as material to be
deleted).  Two words match each other, with no adjustment in the
score, only if they share a major category.

Once all matches have been completed for all possible pairs of root
sequences, the best one(s) are selected. Higher-scoring matches are
preferred, with those involving the deletion of fewer words being
favoured when scores are equal. If there are non-overlapping repairs
(e.g. ``I want to go from from Boston to San San Francisco'') then
the best options for both are accepted.

In the example above, the best path is for the forward match. It
consists simply of recognizing ``no'' as a repair indicator and not
progressing the second pointer at all. This gives a reparandum of
``from Boston no'' and a repair of ``from Denver to Boston'' with a
total score of three.

On the main training corpus of 4615 reference sentences used during
the project, the repair mechanism suggested corrections for 135
sentences. As far as could be determined by inspection of the word
string alone, 89 of these actually were repairs and 41 were not, with
the status of five being impossible to determine without reference to
prosody. The subsequent behaviour of the system for the 130 sentences
whose status was clear was as shown in Table~\ref{tbl:repairs}.
Correct decisions are shown in bold type.

\begin{figure}
\begin{center}
\begin{tabular}{|l|c|c|} \hline
	   	  & Actual 	& False \\
	   	  & repairs	& alarms \\ \hline
No QLF found 	  & 10		& 4 \\  \hline
Right repair chosen   & {\bf 77}	& - \\ \hline
Wrong repair chosen   & 1	& 2 \\ \hline
Non-repair chosen & 1		& {\bf 35} \\ \hline  \hline
Total		  & 89		& 41 \\ \hline
\end{tabular}
\caption{Decisions on possible repairs}
\label{tbl:repairs}
\end{center}
\end{figure}

Restricting attention to sentences for which some QLFs were found, of
the 79 sentences involving repairs for which a QLF for a repaired
version was chosen, the repaired string was correct, or as plausible
as any other choice, in 77 cases. In the other two cases, a wrong
repair, and no repair, were selected respectively. When no repair was
actually present, the preferred QLF was for the unrepaired version in
all but 2 of 37 cases.  Thus the repair mechanism caused 77 sentences
to receive an analysis for the correct string where this would not
otherwise have happened, and caused 2 sentences to receive a bogus
interpretation when they would not otherwise have received one.  In
other words, it increased coverage on the training set by (77-2)/4615,
or 1.6\%.

Of course, performance on reference versions (corresponding to perfect
speech recognition) of training sentences is likely not to be a good
indicator of performance on errorful recognizer outputs for unseen
sentences; and in fact, applying the current repair mechanism to such
outputs does tend to result in the acceptance of noticeably more bogus
repairs, nearly all arising from incorrect sentence hypotheses. As
already indicated, this is quite undesirable.

However, many if not most errors of this type are due to the fact that
the repair mechanism is being applied to a qualitatively different
kind of data from that used to guide its design. We are encouraged by
the fact that, for the reference sentences, a relatively simple repair
{\it suggestion} algorithm can lead to such accurate decisions on the
validity of the repair by the much more sophisticated subsequent
language processing (only 4 wrong choices of string out of 116 cases
where a choice was made). Further work will involve redesigning the
algorithm, and probably training it automatically, to handle the kinds
of output characteristic of the recognizer. As Section
4 below will argue more fully, training language
processing decisions on typical recognizer behaviours rather than only
on reference sentences can enhance decision-making considerably.

\section*{\large \bf \centerline{3. GRAMMAR SPECIALIZATION} \newline
\centerline{FOR FAST PARSING}}

Language models used in the context of speech recognition are normally
some variety of finite-state grammar. Bigram grammars are probably
still the most popular choice, and one is used by the version of
DECIPHER incorporated in SLT. Trigram or higher N-gram models and
stochastic context-free grammars are also fairly common. The
advantages of finite-state models are well-known: they are fast,
robust, and easy to train. The disadvantages are also clear: viewed as
grammatical formalisms, they are insufficiently expressive to capture
many important types of linguistic regularities, and so although they
are useful in the non-final stages of the progressive search task,
they are not adequate for the final stage, nor indeed for constructing
a sufficiently rich semantic representation to support translation.

However, use of more powerful and expressive grammar formalisms tends
to be impractical due to the excessively slow processing times
associated with most known parsing algorithms. This would be
especially problematical in the SLT system when the language analysis
carried out by the CLE counts as a single, final stage of progressive
search, so that many possibilities are considered before any are ruled
out.

In the language analysis part of the SLT system, we have therefore
implemented what we think is an interesting compromise between the
opposing positions of fast finite-state language models and general
linguistically-motivated grammars. The bulk of this work (most of
which has carried out in collaboration with Christer Samuelsson of
SICS, Stockholm) has been reported elsewhere (Rayner, 1988; Rayner and
Samuelsson, 1990; Samuelsson and Rayner, 1991; Samuelsson, 1994). We
content ourselves here with a brief summary relating it to the themes
of the present paper.

We start with a general, linguistically motivated grammar, which has
been given enough specialized vocabulary to have good domain coverage.
In the SLT project, we used the CLE grammar for English (Alshawi,
1992, chapters 4 and 5; Agn\"{a}s {\it et al}, 1994, chapter 7), but
the techniques do not make any special use of its peculiarities, and
would be applicable to any general unification-based phrase-structure
grammar. The key point is that the general grammar is unsuitable for
the language modelling task because it is {\it over}-general; in
particular, there is no need in the context of a normal spoken
language domain to have a fully recursive grammar.

We {\it specialize} the grammar to the domain by first using it to
parse a substantial corpus of examples (in the concrete experiments
carried out, we used a set of about 5000 ATIS sentences). We then
extract a much simpler grammar from the original one by cutting up the
analysis trees from the parsed corpus into sub-trees, where each
sub-tree corresponds to a linguistic ``chunk'' or unit; we used only
four chunk types (utterance, noun phrase, non-recursive noun-phrase
and preposition phrase), compared to about twenty-five different
phrase types in the original grammar. The rules contained in each
sub-tree are then ``collapsed'' into a single rule for the appropriate
chunk-type, using the so-called Explanation-Based Learning algorithm
(van Harmelen and Bundy, 1988; Hirsh, 1987). With a suitable choice of
chunk-types, we can produce a specialized grammar whose rules
correspond to chunk patterns occurring in the training corpus.

By construction, the specialized grammar has strictly less coverage on
the domain than the original one. Our experiments suggest, however,
that given a substantial training corpus the loss of coverage is on
the order of a few percent at most. This loss of coverage is more than
counterbalanced by the greatly simplified structure of the specialized
grammar, which can be parsed nearly two orders of magnitude more
quickly than the general one, using an LR parsing algorithm
(Samuelsson, 1994). The gain in speed is due to the fact that the
grammar, after specialization, is nearly finite-state; we have in
effect automatically squeezed a general grammar into a finite-state
format, after cutting off the few pieces that refuse to fit.

Apart from the enormous gain in speed, it is also worth noting that
the specialized grammar is less ambiguous than the general one; for a
given sentence, it normally produces substantially fewer different
analyses. This implies that the task of identifying a correct analysis
becomes correspondingly simpler. The ``preference component''
described in the next section has less work to do, and makes incorrect
choices less often. In practice, we have discovered that this extra
accuracy more or less cancels out the loss of grammatical coverage;
the few sentences outside specialized grammar coverage tend to be so
complex and ambiguous that there is a high chance of an incorrect
analysis being preferred.

\section*{\large \bf \centerline{4. DISAMBIGUATION}}

Once zero or more QLFs have been produced for each of the original and
repaired sentence hypotheses in the N-best list, the preference
component of the CLE has the task of selecting the most appropriate
one for translation.  It does this by assigning a score to each QLF
and selecting the highest-scoring one, as we will now describe.
A full account is given in Alshawi and Carter (1994).

\subsection*{\large \bf \centerline{4.1 Preference Functions}}

The score assigned to a QLF is a scaled linear sum of the scores
returned by a set of about twenty individual {\it preference
functions}.  Preference functions are of three types.

\begin{itemize}

\item Firstly, there
is a {\it speech} function which simply returns the acoustic score for
the sentence hypothesis that gave rise to the QLF (or a default low
score if the hypothesis was suggested by the repair algorithm).

\item Secondly, {\it structural} functions examine some aspect of the
overall shape of the QLF.  Typically, the number of occurrences of
some relatively unlikely type of grammatical construction is counted,
so that readings which contain instances of it can be penalized
relative to those that do not.

\item Thirdly, {\it combining} functions
collect instances of linguistic objects such as: N-grams in the
underlying word string; the syntax rules used to create the QLF; and
triples of the form $(H_1,R,H_2)$, where $H_1$ and $H_2$ are the head
predicates of QLF substructures representing words or phrases in the
sentence and $R$ indicates the relationship (e.g.\ a preposition or an
argument position) between them. Semantic classes are used to group
place names, numbers and other sets with similar distributions.  For
example, the set of triples for the correct analysis of ``Show me the
flights to Boston'' includes these:
\begin{verbatim}
   (show_CauseToSee,3,flight)
   (flight,to,*place)
\end{verbatim}
the second of which indicates the attachment of ``to Boston'' to
``flights'' rather than to ``show''.  The combining function
calculates, by addition or averaging, a score for the QLF based on the
scores for the individual objects. The objects in turn take their
scores from the pattern of their occurrence in correct and incorrect
QLFs observed in training on recognizer outputs on a corpus for the
domain in question.  Roughly, an object score is intended to be an
estimate of the log probability that a QLF from which the object
arises is the correct one.

\end{itemize}

\subsection*{\large \bf \centerline{4.2 Scaling Factors}}

The scaling factors used to derive a single summed score for a QLF
from the scores returned for that QLF by the various preference
functions are also trained automatically in order to maximize of the
chances of the highest-scoring QLF being correct. Scaling factor
training has two phases.

The first phase makes use of a measure of the similarity between each
QLF for a sentence and the correct QLF (selected in advance by
interaction with a developer) for that sentence.  This measure is
sensitive to differences both in the underlying word sequences and in
the groupings of the words into phrases by the QLFs.  Linear (least
squares) optimization is carried out to find the scaling factor values
that make the preference scores for QLFs resemble the similarity
measures as closely as possible. This is an analytic process that can
be carried out fairly quickly. However, its objective function, that
of modelling similarity to the correct QLF, is only approximately
related to the behaviour we want, that of ensuring that the correct
QLF is placed first in the preference ordering, regardless of the
scores of incorrect QLFs relative to each other.

In the second phase, therefore, scaling factors are adjusted
iteratively to increase the number of training sentences for which the
correct QLF gains the highest score; that is, attention is focused on
selecting correctly among the few most plausible QLFs, and not on
predicting the scores of clearly implausible ones, whose relative
merits are unimportant. Since this task is non-linear, it is fairly
computationally intensive, and may only find a local optimum, so that
the first, linear phase is essential to find a good starting point for
it.

After scaling in this way, the preference functions are able to select
the correct QLF (as judged by an expert) in 90 to 95\% of cases when
trained on four fifths of a corpus of the reference versions of 4092
within-domain, within-coverage ATIS sentences of up to 15 words in
length and tested on the other one fifth, with each one fifth being
held out in turn for testing. This result is for the QLFs produced
with a version of the grammar that had not undergone the
specialization process described in Section 3 above. The figure would
be still higher if only the smaller number of QLFs arising from the
specialized grammar were compared. Thus, as remarked earlier, the
tendency of grammar specialization to reduce coverage slightly is
largely offset by the fact that, for sentences that are still in
coverage, fewer erroneous QLFs are produced which may be preferred
over the correct one.

\subsection*{\large \bf \centerline{4.3 A Comparison}}

To appreciate the importance of some of the points in the above
description, it is instructive to compare the process described above
with the somewhat simpler training procedure used in an earlier
version of the system. For clarity, we will call the earlier version
SLT-0, and current version, implementing the above procedure, SLT-1.
SLT-0 lost some accuracy because in it, the various scores and scaling
factors were optimized for tasks related to, but not identical to,
that encountered at run-time.

The first difference is that in SLT-0, the linguistic objects used by
some of the combining metrics were scored not by comparing good and
bad QLFs but on the basis only of their frequency of occurrence in
good QLFs.  As we will see in the next section, this is suboptimal,
because an object is not a good predictor of correctness simply
because it occurs frequently in good QLFs; it may occur just as often
in bad ones.

SLT-0's second drawback was that training with respect to the corpus
was decoupled from training with respect to the speech recognizer.
That is, object scores and all the non-speech scaling factors were
calculated by looking only at QLFs for the reference versions of
corpus sentences, and not at recognizer outputs.  The scaling factor
for the speech function was then found by trial and error on a
separate training corpus. Thus SLT-0 had no opportunity to adapt to
and compensate for typical recognizer errors.

QLF selection accuracy turned out in fact to be relatively insensitive
to the value of the acoustic factor, which can be doubled or halved
without noticeable effect. However, the lack of training on incorrect
sentence hypotheses was a more serious drawback. There are syntactic
and semantic patterns which seldom occur in analyses of correct
sentence hypotheses and therefore were not assigned very large scores,
but which often crop up as a consequence of certain kinds of
recognizer error.  A known example of this behaviour is number
disagreement between subject and predicate in a sentence hypothesis
with main verb ``be'', for example ``What is the first flights to
Boston?''. This is grammatically possible but most unlikely to be
correct, and usually indicates that the head noun of the predicate
phrase has been recognized with the wrong number: in the example, the
word spoken would actually have been ``flight''. There are also
examples of semantic triples, and perhaps also syntax rules, which
likewise tend to characterize analyses of wrong hypotheses but which,
for that very reason, are not observed when training only on correct
word strings.  It is not sufficient to finesse this problem by
penalizing objects only observed infrequently in training on reference
sentences, because there is no {\it a priori} way of knowing whether
such an object, when encountered at run time, indicates a recognizer
error or just an unusual, but genuine, form of words. In the next
section, we focus in more detail on this problem and how it is
is overcome.

\subsection*{\large \bf \centerline{4.4 Tuning to the N-Best Task}}

The deficiencies just described for SLT-0 had the effect that
selecting a sentence hypothesis using the trained combination of
speech, structural and combining preference functions only yielded a
2\% increase in sentence accuracy (as measured on a 1000-sentence
unseen test set) over using the speech score alone.  This figure is in
a sense misleadingly pessimistic, since we are interested in
translation rather than recognition {\it per se}, and the combined
functions always select a hypothesis for which a QLF, and therefore
potentially a translation, is found, whereas the recognizer alone
sometimes prefers an unanalysable string, which even if correct will
not be translated. Nevertheless, it seemed likely that introducing
linguistic factors, if done optimally, should improve sentence
accuracy by more than a couple of per cent.

We therefore carried out some experiments (reported in full in Rayner
{\it et al}, 1994) in which several preference functions were trained
on N-best data as in SLT-1, but with sentence hypothesis selection,
rather than QLF selection, as the objective. The value of N was chosen
to be 10, rather than 5 as in the run-time system. The preference
functions used were:
\begin{itemize}
\item The speech function, returning the recognizer score.
\item Two structural functions: one which returned 1 if any QLFs were
found for the sentence using the specialized grammar, and otherwise 0;
and one which returned 1 if the best QLF for the string (as judged by
the existing preference module) contained a subject-predicate number
mismatch, and otherwise 0.
\item Two combining functions: one for grammar rules used in the best
QLF for the string, and one for the semantic triples for that QLF.
\end{itemize}
We found that over the 1000-sentence test set, the optimized
combination of functions selected the correct hypothesis 70.5\% of the
time, compared to a maximum possible 84.2\% where the correct
hypothesis occurred at all in the 10-best list, a score of 66.3\% for
the speech function alone, and a score of 67.8\% for the more
traditional approach of selecting the first hypothesis in the
recognizer list that received a parse. Thus the optimized combination
gave an absolute improvement over the speech function alone of 4.2\%,
double the corresponding figure for SLT-0.  For sentences of 12 words
and under, the improvement was 5.6\%. These sentences showed a larger
improvement because they were more likely to be analysable by the CLE;
if no QLFs are produced for any hypothesis, the linguistic functions
have no contribution to make. Because of this drawback, it turned out
that N-gram combining functions for N=1 to 4, which can be applied
even when no QLFs are produced, were slightly more powerful in
combination with the speech function than the CLE-based functions
were, although for 12-word sentences and the acceptable variant
criterion, no difference was apparent. Not surprisingly, when N-gram
knowledge sources were added, a still better result, 73.7\%, was
obtained.

We concluded from these results that it is well worth training
linguistic functions in this way. One further possible improvement is
that for sentence recognition (although probably not for translation,
because of the risk of errors), it would also be desirable to derive
QLF analyses of parts of a sentence when no full analysis could be
found; this would allow linguistic functions always to make some
contribution, even if only an imperfect one, and would improve
accuracy on utterances for which no hypothesis was perfectly correct
and those which included constructions outside the coverage of the
grammar.

\section*{\large \bf \centerline{5. SUMMARY AND} \newline
\centerline{CONCLUSIONS}}

We have described the ways in which language analysis in SLT makes
intelligent use of the N-best hypothesis list delivered by the speech
recognizer, implementing the final stage of progressive search by
avoiding nearly all hard decisions about word identities or sentence
meanings until all available linguistic knowledge has been applied.
That is, the CLE creates its whole search space before pruning away
any of it.  Thus alternative QLF analyses for the same recognizer
hypothesis, for different recognizer hypotheses, and for repaired as
well as unrepaired versions of hypotheses are all constructed and
compared in a uniform way. The use of an automatically tuned grammar
and associated fast parser makes this generate-and-test process
acceptably fast (typically a few seconds per speech hypothesis on a
SPARCstation 10) by eliminating many impossible search paths and some
possible but unlikely ones.

It would be possible to speed up the system further by parallelizing
it. Each recognizer hypothesis could be analysed separately, and the
highest-scoring QLF (if any) resulting from it returned for a final
choice to be made.

The unattainable ideal in any search problem is for the search space
constructed to consist only of the correct solution, or of solutions
that are equally likely to be correct. We approximate this ideal in
the speech understanding task by training and selecting grammar rules
(the objects that generate possible solutions) on human-transcribed
reference material, so that, as far as possible, correct solutions
will fall within the search space and incorrect ones will fall outside
it. In practice, of course, by no means all incorrect solutions will
be excluded in this way; so we train preference functions on
recognizer and language analysis output, to maximize our chances of
distinguishing correct from incorrect solutions, whatever stage of
processing they arise from.

In Section 4.4 above we gave performance details for speech and
language analysis. Sentence recognition accuracy using optimized
speech (DECIPHER) and language (CLE and N-gram) information on unseen
ATIS data is 73.7\%.  Full details of the performance of an earlier
version of the full system (roughly SLT-0) are given by Rayner {\it et
al}, 1993. Briefly, however, for sentences within the ATIS domain and
up to twelve words in length, if a correct speech hypothesis is
selected then a Swedish translation is produced on about three
occasions in four, and 90\% of those translations are acceptable.  The
remaining 10\% can nearly all be clearly identified by the hearer as
errors because they are ungrammatical or unnatural; divergences
in meaning, which might lead to more serious forms of dialogue
failure, are extremely rare.

\section*{\centerline{\large \bf ACKNOWLEDGEMENTS}}

This paper appears in the Proceedings of TWLT-8, the Twente Workshop
on Language Technology, 1994, edited by L.~Boves and A.~Nijholt.

The bulk of the work described here was done on a project funded by
Telia Research AB. The partners were SRI International, Menlo Park
(speech recognition); SRI International, Cambridge (language
processing software and English grammar); the Swedish Institute for
Computer Science (English-Swedish transfer rules, Swedish grammar and
fast parser for specialized grammar), and Telia Research (speech
synthesis).  Adaptation to French was carried out by ISSCO, Geneva, in
collaboration with SRI Cambridge.

\section*{\centerline{\large \bf REFERENCES}}

Agn\"{a}s, M-S., and 17 others (1994). {\em Spoken Language
Translator: First Year Report}. Joint report by SRI International
(Cambridge) and SICS. Order from \verb!preben@sics.se!.

\noindent
Alshawi, Hiyan, editor (1992). {\em The Core Language Engine}.  The
MIT Press, Cambridge, Massachusetts.

\noindent
Alshawi, Hiyan, and David Carter (1994). ``Training and Scaling
Preference Functions for Disambiguation''. {\it Computational
Linguistics}, 20:4.

\noindent
Alshawi, Hiyan, and Richard Crouch (1992).  ``Monotonic Semantic
Interpretation''.  In {\em Proceedings of 30th Annual Meeting of the
Association for Computational Linguistics}, pp.~32--39, Newark,
Delaware.

\noindent
Bear, J., J.~Dowding, and E.~Shriberg (1992).  ``Integrating
multiple knowledge sources for the detection and correction of repairs
in human-computer dialog''.  In {\em Proceedings of 30th Annual
Meeting of the Association for Computational Linguistics}, pp.~56--63,
Newark, Delaware.

\noindent
van Harmelen, Frank, and Alan Bundy (1988). ``Explanation-Based
Generalization = Partial Evaluation'' (Research Note), {\it Artificial
Intelligence 36\/}, pp.~401--412.

\noindent
Hirsh, Haym (1987).  ``Explanation-Based Generalization in a
Logic-Programming Environment'', {\it Proceedings of the Tenth
International Joint Conference on Artificial Intelligence\/}, Milan,
pp.~221--227.

\noindent
Murveit, Hy, John Butzberger, Vassilios Digalakis, and Mitch Weintraub
(1993).  ``Large Vocabulary Dictation using SRI's DECIPHER(TM) Speech
Recognition System: Progressive Search Techniques''.  In {\em
Proceedings of the IEEE International Conference on Acoustics, Speech
and Signal Processing}, volume~2, pp.~319--322, Minneapolis,
Minnesota.

\noindent
Murveit, Hy, John Butzberger, and Mitch Weintraub (1991). ``Speech
Recognition in SRI's Resource Management and ATIS Systems''.  In {\em
Proceedings of the 4th Speech and Natural Language Workshop}.  DARPA,
Morgan Kaufmann.

\noindent
Nakatani, C., and J.\ Hirschberg (1993).  ``A speech-first model of
repair detection and correction''.  In {\em Proceedings of 31th Annual
Meeting of the Association for Computational Linguistics}, pp.~46--53,
Columbus, Ohio.

\noindent
Rayner,~M. (1988). ``Applying Explanation-Based Generalization to
Natural-Language Processing''. {\it Proceedings of the Conference on
Fifth Generation Computer Systems}, Tokyo.

\noindent
Rayner,~M., and C.~Samuelsson (1990). ``Using Explanation-Based
Learning to Increase Performance in a Large-Scale NL Query
Interface''.  {\it Proceedings of the 3rd DARPA Speech and Natural
Language Workshop}, Hidden Valley.

\noindent
Rayner,~M., and 11 others (1993). ``Spoken Language Translation with
Mid-90's Technology: a Case Study''. {\it Proceedings of
Eurospeech-93}, Berlin.

\noindent
Rayner,~M., D.~Carter, V.~Digalakis and P.~Price (1994). ``Combining
Knowledge Sources to Reorder N-Best Speech Hypothesis Lists''.  {\it
Proceedings of the 1994 ARPA Workshop on Human Language Technology},
Princeton.

\noindent
Samuelsson,~C., and M.~Rayner (1991). ``Quantitative Evaluation of
Explanation-Based Learning as a Tuning Tool for a Large-Scale Natural
Language System''. {\it Proceedings of the 12th International Joint
Conference on Artificial Intelligence}, Sydney.

\noindent
Samuelsson,~C. (1994). {\it Fast Natural Language Parsing Using
Explanation-Based Learning}, PhD thesis, Royal Institute of
Technology, Stockholm.

\end{document}